\documentclass[twoside]{article}
\usepackage[accepted]{aistats2016}
\usepackage[authoryear]{natbib}
\usepackage{hyperref}
\usepackage{url}
\usepackage{graphicx} % more modern
\usepackage{subfigure} 
\usepackage{algorithm}
\usepackage{algorithmic}
\usepackage{amsmath}[11pt]
\usepackage{amsmath, amsthm, amssymb}
\usepackage{amsthm}
\usepackage{amsfonts}[11pt]
\usepackage{mathtools}
\usepackage{bbm}

\newcommand{\bc}{\begin{center}}
\newcommand{\ec}{\end{center}}
\newcommand{\bq}{\begin{quote}}
\newcommand{\eq}{\end{quote}}

\newcommand{\be}{\begin{equation}}
\newcommand{\ee}{\end{equation}}
\newcommand{\beqa}{\begin{eqnarray*}}
\newcommand{\eeqa}{\end{eqnarray*}}
\newcommand{\beqn}{\begin{eqnarray}}
\newcommand{\eeqn}{\end{eqnarray}}
\newcommand{\bbibl}{\begin{thebibliography}{99}}
\newcommand{\ebibl}{\end{thebibliography}}

\newcommand{\ba}{\begin{array}}
\newcommand{\ea}{\end{array}}

\newcommand\inner[1]{\left\langle {#1} \right\rangle}

\newtheorem{lemma}{Lemma}

\newtheorem{theorem}{Theorem}[section]

\DeclareMathOperator*{\argsort}{argsort}

\DeclareMathOperator*{\argmin}{argmin}

\newcommand\reals{\mathbb{R}}
\newcommand{\E}{\mathbb{E}}
\newcommand*{\permcomb}[4][0mu]{{{}^{#3}\mkern#1#2_{#4}}}
\newcommand*{\perm}[1][-3mu]{\permcomb[#1]{P}}
\newcommand\listnet{\phi_{\mathrm{LN}}}
\newcommand\sdcg{\phi_{\mathrm{SD}}}

\begin{document}



\twocolumn[

\aistatstitle{Online Learning to Rank with Feedback at the Top}

\aistatsauthor{ Sougata Chaudhuri \And Ambuj Tewari}

\aistatsaddress{ University of Michigan, Ann Arbor  \And University of Michigan, Ann Arbor } ]

\begin{abstract}
We consider an online learning to rank setting in which, at each round, an oblivious adversary generates a list of $m$ documents, pertaining to a query, and the learner produces scores to rank the documents. The adversary then generates a relevance vector and the learner updates its ranker according to the feedback received. We consider the setting where the feedback is restricted to be the relevance levels of only the top $k$ documents in the ranked list for $k \ll m$. However, the performance of learner is judged based on the unrevealed full relevance vectors, using an appropriate learning to rank loss function. We develop efficient algorithms for well known losses in the pointwise, pairwise and listwise families. We also prove that no online algorithm can have sublinear regret, with top-1 feedback, for any loss that is calibrated with respect to NDCG. We apply our algorithms on benchmark datasets demonstrating efficient online learning of a ranking function from highly restricted feedback. 
\end{abstract}

\section{Introduction}
In learning to rank for information retrieval, the objective is to rank lists of documents, pertaining to different queries, so that the documents that are more relevant to a query are ranked above those that are less relevant. Most learning to rank methods are based on supervised \emph{batch} learning, i.e., rankers are trained on batch data consisting of instances and labels \citep{liu2011learning}. The instances are lists of documents, pertaining to different queries, and labels are in the form of relevance vectors. The accuracy of a ranked list, in comparison to the actual relevance of the documents, is measured by various ranking measures, such as NDCG, AP and ERR.

Collecting reliable training data can be expensive and time consuming. In certain applications, such as deploying a new web app or developing a custom search engine, collecting large amount of training data might not be possible at all \citep{sanderson2010test}. Moreover, a ranker trained from batch data might not be able to satisfy changing user needs and preferences.
Recent research has focused on online learning of ranking systems, where a ranker is updated on the fly. One direction of work deploys models which learn from implicit feedback inferred from user clicks on ranked lists \citep{hofmann2013balancing,yue2009interactively}. However, there are some potential drawbacks in learning from user clicks. It is possible that the displayed items might not be clickable, such as in certain mobile apps. Moreover, a clicked item might not actually be relevant to the user and there is also the problem of bias towards top ranked items in inferring feedback from user clicks \citep{joachims2002}. Another direction of work deploys models which learn optimal ranking of a fixed list of items, for diverse user preferences \citep{radlinski2008learning,chaudhuri2015}. Specifically, the latter work assumes that a user generates a full relevance vector for the entire ranked list of items but gives feedback only on the top ranked item. Motivation for this feedback model comes from considerations of user burden constraints (users will feel burdensome to provide careful feedback on all items) and privacy concerns (users will be unwilling to provide feedback on all items if they are about sensitive issues such as medical conditions). However, the requirement of having a fixed set of items to rank severely limits the practical applicability of this line of work.

Our work extends the work of \cite{chaudhuri2015}, by combining query-level ranking, in an online manner, with explicit but restricted feedback. We formalize the problem as an online game played over $T$ rounds, between a learner and an \emph{oblivious} adversary. At each round, the adversary generates a document list of length $m$, pertaining to a query. The learner sees the list and produces a real valued score vector to rank the documents. We assume that the ranking is generated by sorting the score vector in descending order of its entries. The adversary then generates a relevance vector but the learner gets to see the relevance of only the top-$k$ items of the ranked list, where $k \ll m$ is a small constant, like $1$ or $2$. The learner's loss in each round, based on the learner's score vector and the \emph{full} relevance vector, is measured by some continuous learning to rank loss function. We focus on continuous surrogates losses, e.g., the cross entropy surrogate in ListNet \citep{cao2007learning} and hinge surrogate in RankSVM \citep{joachims2002}, instead of discontinuous ranking measures like NDCG, AP, or ERR because the latter lead to intractable optimization problems. We note that the top-$k$ feedback model is distinct from the full and bandit feedback models since neither the full relevance vector nor the loss at end of each round is  revealed to the learner. Technically, the problem is an instance of \emph{partial monitoring} \citep{cesa2006, bartok2013}, \emph{extended to a setting with side information} (documents list) and an \emph{infinite set of learner's moves} (all real valued score vectors). For such an extension of partial monitoring there exists no generic theoretical or algorithmic framework to the best of our knowledge. 

We make two main contributions in this paper. First, we propose a general, efficient algorithm for online learning to rank with top-$k$ feedback and show that it works in conjunction with a number of ranking surrogates. We characterize the minimum feedback required, i.e., the value of $k$, for the algorithm to work with a particular surrogate by formally relating the feedback mechanism with the structure of the surrogates. We then apply our general techniques to three convex ranking surrogates and one non-convex surrogate. The convex surrogates considered are from three major learning to ranking methods: squared loss from a \emph{pointwise} method \citep{cossock2008statistical}, hinge loss used in the \emph{pairwise} RankSVM \citep{joachims2002} method, and (modified) cross-entropy surrogate used in the \emph{listwise} ListNet \citep{cao2007learning} method. The non-convex surrogate considered is the SmoothDCG surrogate \citep{chapelle2010gradient}. For the three convex surrogates, we establish an $O(T^{2/3})$ regret bound.

The convex surrogates we mentioned above are widely used but are known to fail to be calibrated with respect to NDCG \citep{ravikumar2011ndcg}. Our second contribution is to show that for the entire class of NDCG calibrated surrogates, no online algorithm can have sublinear (in $T$) regret with top-1 feedback, i.e., the minimax regret of an online game for any NDCG calibrated surrogate is $\Omega(T)$. The proof for this rather surprising result is non-trivial and relies on exploiting a connection between the construction of optimal adversary strategies for hopeless \emph{finite action} partial monitoring games \citep{piccolboni2001discrete} and the structure of NDCG calibrated surrogates. We only focus on NDCG calibrated surrogates for the \emph{impossibility} results since no (convex) surrogate can be calibrated for AP and ERR \citep{calauzenes2012non}. This impossibility result is not only the first of its kind in online ranking with top-$1$ feedback but it also the first such result for a natural partial monitoring problem with side information when the learner's action space is infinite. Note, however, that there does exist work on partial monitoring problems with continuous learner actions, but without side information \citep{kleinberg2003value,cesa2006}, and vice versa \citep{bartok2012partial,gentile2014multilabel}.

We apply our algorithms on benchmark ranking datasets, demonstrating the ability to efficiently learn a ranking function in an online fashion, from highly restricted feedback.
 %We propose an algorithm for online learning to rank with top-$k$ feedback that incurs low regret according to any of a number of convex surrogates used in established learning to rank methods. Our algorithm requires the construction of unbiased estimators of the gradient of ranking surrogates with respect to the score vector. We characterize the general construction process of unbiased estimators from partial feedback and construct estimators for three major convex surrogates and one non-convex surrogate. The convex surrogates considered are from three major learning to ranking methods: squared loss from \emph{pointwise} method \citep{cossock2008statistical}, hinge loss based surrogate used in the \emph{pairwise} RankSVM \citep{joachims2002} method, and (modified) cross-entropy surrogate used in the \emph{listwise} ListNet \citep{cao2007learning} method. The non-convex surrogate considered is the SmoothDCG surrogate \citep{chapelle2010gradient}. For convex surrogates, we also establish formal regret guarantee, which scales as $O(T^{2/3})$ for all the three convex surrogates. We apply our algorithms on benchmark ranking datasets, demonstrating the ability to efficiently learn a ranking function in an online fashion, from highly restricted feedback. 

\section{Preliminaries}
\label{prelim}
In learning to rank, an instance is a matrix $X \in \mathbb{R}^{m \times d}$, consisting of a list of $m$ documents, each represented as a feature vector in $\mathbb{R}^d$, with each list pertaining to a single query. 
The supervision is in form of a relevance vector $R= \{0,1,\ldots,n\}^m$, representing relevance of each document to the query. If $n=1$, the relevance vector is binary graded. For $n>1$, relevance vector is multi-graded. $X_{i:}$ denotes $i$th row of $X$ and $R_i$ denotes $i$th component of $R$. \emph{The subscript $t$ is exclusively used to denote time $t$}. Thus, $R_t$ denotes relevance vector generated at time $t$ and $R_{t,i}$ denotes $i$th component of $R_t$. We assume feature vectors representing documents are bounded by $R_D$ in $\ell_2$ norm.

Documents are ranked by a ranking function. The prevalent technique is to represent a ranking function as a scoring function and get ranking by sorting scores in descending order. A linear scoring function produces score vector as $f_{w}(X)= Xw= s^w \in \mathbb{R}^m$, with $w \in \mathbb{R}^d$. Here, $s^w_i$ represents score of $i$th document ($s^w$ points to score $s$ being generated by using parameter $w$). We assume that ranking parameter space is bounded in $\ell_2$ norm, i.e, $\|w\|_2 \le U$, $\forall \ w$. $\pi_s = \argsort(s)$ is the permutation induced by sorting score vector $s$ in descending order. A permutation $\pi$ gives a mapping from ranks to documents and $\pi^{-1}$ gives a mapping from documents to ranks. Thus, $\pi(i)=j$ means document $j$ is placed at position $i$ in $\pi$ while $\pi^{-1}(i)=j$ means document $i$ is placed at position $j$. $S_m$ denote the set of $m!$ different permutations of $[m]$ where $[m]= \{1,2\ldots,m\}$.

Various ranking measures, like NDCG and AP, judge the quality of a ranking function, by comparing the ranked lists produced by the ranking function and the relevance vector, respectively.  Formally, NDCG, cut off at $k \le m$ for a query with $m$ documents, with relevance vector $R$ and score vector $s$ induced by a ranking function, is defined as follows: $\text{NDCG}_k(s,R) = \frac{1}{Z_k(R)}\sum_{i=1}^k G(R_{\pi_s(i)})D(i)$.
%\begin{equation}
%\label{eq:NDCG}
%\begin{split}
%NDCG(s,R)= \frac{1}{Z(R)}\sum_{i=1}^m G(R_i)D(\pi^{-1}_s(i)) 
%\end{split}
%\end{equation}
Shorthand representation of $\text{NDCG}_k(s,R)$ is $\text{NDCG}_k$. Here, $G(r)= 2^r -1$, $D(i)= \frac{1}{\log_2{(i+1)}}$, $Z_k(R)= \underset {\pi \in S_m}{\max}\sum_{i=1}^k G(R_{\pi(i)})D(i)$. $\pi_s= \argsort(s)$ is the permutation induced by score vector $s$ in descending order. Since optimization of the discontinuous ranking measures is an NP-hard problem, most ranking methods are based on minimizing \emph{surrogate} losses, which can be optimized more efficiently. A surrogate $\phi$ takes in a score vector $s$ and relevance vector $R$ and produces a real number, i.e., $\phi: \mathbb{R}^m \times \{0,1,\ldots,n \}^m \mapsto \mathbb{R}$. $\phi(\cdot,\cdot)$ is said to be convex if it is convex in its first argument, for any value of the second argument. The ranking surrogates are designed in such a way that the score vector which minimizes the surrogate, induces a ranking which minimizes the target ranking measures.  
%We use two performance measures in our paper: (N)DCG ((Normalized) Discounted Cumulative Gain) and AP (Average Precision). (N)DCG, for a set of $m$ documents, with relevance vector $R$ and score vector $s$, is defined as follows : $\text{(N)DCG}(\pi_s,R)= \frac{1}{Z(R)}\sum_{i=1}^{m} G(R_{\pi_s(i)})D(i)$, where $G(r)= 2^r -1$, $D(i)= \frac{1}{\log_2{(i+1)}}$, and normalization factor $Z(R)= \underset {\pi}{\max}\sum_{i=1}^{m} G(R_{\pi(i)})D(i)$. AP is defined only for binary relevance vector: $\text{AP}(\argsort(s),R) = \frac{1}{r} \sum_{j:R_{\pi_s(j)}=1} \frac{\sum_{i \le j} \mathbbm{1}[R_{\pi_s^(i)}=1]}{j}$, where $r= \|R\|_1$ is the total number of relevant documents.

\section{Problem Setting and Learning to Rank Algorithm}
{\bf Formal problem setting}: We formalize the problem as a game being played between a learner and an \emph{oblivious} adversary over $T$ rounds. The learner's action set is the uncountably infinite set of score vectors in $\mathbb{R}^m$ and the adversary's action set is all possible relevance vectors, i.e., $(n+1)^m$ possible vectors. At round $t$, the adversary generates a list of documents, represented by a matrix $X_t \in \mathbb{R}^{m \times d}$, pertaining to a query (the document list is considered as side information). The learner receives $X_t$ and produces a score vector $\tilde{s}_t \in \reals^m$. The adversary then generates a relevance vector $R_t$ but only reveals the relevance of top $k$ ranked documents to the learner, where the ranked list is produced by sorting $\tilde{s}_t$. The learner uses the feedback to choose its action for the next round (updates an internal scoring function). The learner suffers a loss as measured in terms of a surrogate $\phi$, i.e, $\phi(\tilde{s}_t,R_t)$. \emph{Note that since the learner's objective is to produce good ranking at every round, learner's performance is measured w.r.t. to entire relevance vector $R_t$ whereas it only gets to see just the top-$k$ entries of $R_t$}. As is standard in online learning setting, the learner's performance is measured in terms of its expected regret: 
%\begin{equation}
$\E \left[\sum_{t=1}^T \phi(\tilde{s}_t,R_t) \right] - \min_{\|w\|_2 \le U} \sum_{t=1}^T \phi(X_tw, R_t)$,
%\end{equation} 
where the expectation is taken w.r.t. to randomization of learner's strategy and $X_tw = s_t^w$ is the score produced by the linear function parameterized by $w$.
% that can be chosen with full knowledge of the $R_t$'s.

\floatstyle{ruled}
\newfloat{algorithm}{htbp}{loa}
\floatname{algorithm}{Algorithm}
\begin{algorithm*}
\caption{Ranking with Top-k Feedback (RTop-kF)}
\label{alg:RTop-kF}
\begin{tabbing}
%tabs \= tabs \= tabs \= tabs kill
1: Exploration parameter $\gamma \in (0,\frac{1}{2})$, learning parameter $\eta>0$, ranking parameter  $w_1=\mathbf{0} \in \mathbb{R}^d$\\
2: {\bf For} \=$t=1$ to $T$ \\
3: \> Receive $X_t$ (document list pertaining to query $q_t$)\\
4: \> Construct score vector $s_t^{w_t}= X_tw_t$ and get permutation $\sigma_t= \argsort(s_t^{w_t})$ \\
5: \> $\mathbb{Q}_t(s) = (1-\gamma)\delta(s-s^{w_t}_t) + \gamma\text{Uniform}([0,1]^m)$ ($\delta$ is the Dirac Delta function). \\
6: \> Sample $\tilde{s}_t \sim \mathbb{Q}_t$ and output the ranked list $\tilde{\sigma}_t = \argsort(\tilde{s}_t)$ \\
   \> (Effectively, it means $\tilde{\sigma}_t$ is drawn from $\mathbb{P}_t(\sigma)= (1-\gamma)\mathbbm{1}(\sigma=\sigma_t) + \frac{\gamma}{m!}$) \\
7: \> Receive relevance feedback on top-$k$ items, i.e., ($R_{t, \tilde{\sigma}_t(1)}, \ldots, R_{t,\tilde{\sigma}_t(k)}$)\\
8: \> Suffer loss $\phi(\tilde{s}_t,R_t)$ (Neither loss nor $R_t$ revealed to learner) \\
9: \> Construct $\tilde{z}_t$, an unbiased estimator of gradient $\nabla_{w=w_t} \phi(X_t w,R_t)$, from  top-$k$ feedback.\\ 
10: \> Update $w= w_t - \eta \tilde{z}_t$ \\
11: \> $w_{t+1}= \min \{1, \frac{\text{U}}{\|w\|_2}\} w$ (Projection into Euclidean ball of radius $U$).\\
12: {\bf End For}
\end{tabbing}
\end{algorithm*} 
%\footnotetext{From the perspective of ranking surrogate $\phi$, the algorithm plays $s_t= s^{w_t}_t$ w.p. $1-\gamma$ (exploitation) and any score vector, selected ``uniformly at random'' (i.e, $s_t= \text{random score vector}$), from $\ell_2$ norm bounded ball in $\mathbb{R}^m$, w.p $\gamma$ (exploration)}.

{\bf Relation between feedback and structure of surrogates:}
Alg.\ \ref{alg:RTop-kF} is our general algorithm for learning a ranking function, online, from partial feedback. 
The key step in Alg.\ \ref{alg:RTop-kF} is the construction of the unbiased estimator $\tilde{z}_t$ of the surrogate gradient $\nabla_{w=w_t} \phi(X_t w, R_t)$. The information present for the construction process, at end of round $t$, is the random score vector $\tilde{s}_t$ (and associated permutation $\tilde{\sigma}_t$) and relevance of top-$k$ items of $\tilde{\sigma}_t$, i.e.,  $\{R_{t, \tilde{\sigma}_t(1)}, \ldots, R_{t,\tilde{\sigma}_t(k)}\}$. Let $\E_{t}\left[\cdot\right]$ be the expectation operator w.r.t. to randomization at round $t$, conditioned on $(w_1,\ldots, w_t)$. Then $\tilde{z}_t$ being an unbiased estimator of gradient of surrogate, w.r.t $w_t$,  means the following: $\E_{t} \left[\tilde{z}_t \right]= \nabla_{w=w_t} \phi(X_tw, R_t)$. We note that conditioned on the past, the score vector $s^{w_t}_t = X_t w_t$ is deterministic. We start with a general result relating feedback to the construction of unbiased estimator of a vector valued function. Let $\mathbb{P}$ denote a probability distribution on $S_m$, i.e, $\sum_{\sigma \in S_m} \mathbb{P}(\sigma)= 1$. For a distinct set of indices $(j_1,j_2,\ldots, j_k)$ $\subseteq$ $[m]$, we denote $p(j_i,j_2,\ldots,j_k)$ as the the sum of probability of permutations whose first $k$ objects match objects $(j_1, \ldots, j_k)$, in order. Formally,
\begin{equation}
\label{eq:shortprob}
\begin{split}
p(j_1,  \ldots, j_k) = 
 \sum\limits_{\pi \in S_m}\mathbb{P}(\pi) \mathbbm{1}(\pi(1)= j_1,\ldots,\pi(k)=j_k) .
\end{split}
\end{equation}
%where, as a reminder, $\pi(i)=j$ means object $j$ is ranked in position $i$ in $\pi$. 
\begin{lemma}
\label{unbiasedestimator}
Let $F: \mathbb{R}^m \mapsto \mathbb{R}^a$ be a vector valued function, where $m\ge 1$, $a\ge 1$. For a fixed $x \in \mathbb{R}^m$, let $k$ entries of $x$ be observed at random. That is, for a fixed probability distribution $\mathbb{P}$ and some random $\sigma \sim \mathbb{P}(S_m)$, observed tuple is $\{\sigma, x_{\sigma(1)}, \ldots, x_{\sigma(k)}\}$. A necessary condition for existence of an unbiased estimator of $F(x)$, that can be constructed from $\{\sigma, x_{\sigma(1)}, \ldots, x_{\sigma(k)}\}$, is that it should be  possible to decompose $F(x)$ over $k$ (or less) coordinates of $x$ at a time. That is, $F(x)$ should have the structure:
\begin{equation}
\label{eq:decoupling}
F(x)= \sum\limits_{(i_1,i_2,\ldots,i_{\ell}) \in \ \perm{m}{\ell}} h_{i_1,i_2,\ldots,i_{\ell}}(x_{i_1}, x_{i_2},\ldots, x_{i_{\ell}}) 
\end{equation}
where $\ell \le k$, $\perm{m}{\ell}$ is $\ell$ permutations of $m$ and $h: \mathbb{R}^{\ell} \mapsto \mathbb{R}^a$ (the subscripts in $h$ is used to simply represent different functions).
Moreover, when $F(x)$ can be written in form of Eq~\ref{eq:decoupling} , with $\ell=k$, an unbiased estimator of $F(x)$, based on $\{\sigma, x_{\sigma(1)}, \ldots, x_{\sigma(k)}\}$, is, 
%\begin{equation}
%\label{eq:unbiasedestimator}
%g(\sigma, x_{\sigma(1)}, \ldots, x_{\sigma(k)})=  \dfrac{h_{\sigma(1),\ldots, \sigma(k)}(x_{\sigma(1)}, \ldots, x_{\sigma(k)})}{p(\sigma(1),\ldots,\sigma(k))} 
%\end{equation}
%where $p(\sigma(1),\ldots, \sigma(k)) = \sum\limits_{\pi \in S_m}\mathbb{P}(\pi) \mathbbm{1}(\pi(1)= \sigma(1),\ldots,\pi(k)=\sigma(k))$ (as in Eq.~\ref{eq:shortprob} )
\begin{equation}
\label{eq:unbiasedestimator}
\begin{split}
g(\sigma, & x_{\sigma(1)}, \ldots, x_{\sigma(k)})=\\
& \dfrac{\sum\limits_{(j_1,j_2,\ldots, j_k) \in S_k}h_{\sigma(j_1),\ldots, \sigma(j_k)}(x_{\sigma(j_1)}, \ldots, x_{\sigma(j_k)})}{\sum\limits_{\substack{(j_1,\ldots, j_k) \in S_k}} p(\sigma(j_1),\ldots,\sigma(j_k))} 
\end{split}
\end{equation}
where $S_k$ is the set of $k!$ permutations of $[$k$]$ and $p(\sigma(1),\ldots, \sigma(k))$ is as in Eq~\ref{eq:shortprob} .
\end{lemma}

{\bf Illustrative Examples:} We provide simple examples to concretely illustrate the abstract functions in Lemma~\ref{unbiasedestimator}. Let $F(\cdot)$ be the identity function, and $x \in \reals^m$.  Thus, $F(x)=x$ and the function decomposes over $k=1$ coordinate of x as follows: $F(x)= \sum_{i=1}^m x_i e_i$, where $e_i \in \mathbb{R}^m$ is the standard basis vector along coordinate $i$. Hence, $h_{i}(x_{i})= x_{i} e_{i}$. Based on top-1 feedback, following is an unbiased estimator of $F(x)$: $g(\sigma, x_{\sigma(1)})= \dfrac{x_{\sigma(1)}e_{\sigma(1)}}{p(\sigma(1))}$,
where $p(\sigma(1))=  \sum\limits_{\pi \in S_m}\mathbb{P}(\pi) \mathbbm{1}(\pi(1)= \sigma(1))$. In another example, let $F: \reals^3 \mapsto \reals^2$ and $x \in \reals^3$. Let $F(x)= [x_1+ x_2; x_2+x_3]^{\top}$. Then the function decomposes over $k=1$ coordinate of $x$ as $F(x)= x_1 e_1 + x_2 (e_1 + e_2) + x_3 e_2$, where $e_i \in \reals^2$. Hence, $h_1 (x_1) = x_1 e_1$, $h_2(x_2)= x_2(e_1 + e_2)$ and $h_3(x_3)= x_3 e_2$. An unbiased estimator based on top-1 feedback is: $g(\sigma, x_{\sigma(1)})= \dfrac{h_{\sigma(1)}(x_{\sigma(1)})}{p(\sigma(1))}$.

\section{Unbiased Estimators of Gradients of Surrogates}
Alg.\ \ref{alg:RTop-kF} can be implemented for any ranking surrogate as long as an unbiased estimator of the gradient can be constructed from the random feedback. We will use techniques from \emph{online convex optimization} to obtain formal regret guarantees. We will thus construct the unbiased estimator of four major ranking surrogates. Three of them are popular \emph{convex} surrogates, one each from the three major learning to rank methods, i.e., \emph{pointwise}, \emph{pairwise} and \emph{listwise} methods. The fourth one is a popular \emph{non-convex} surrogate.

{\bf Shorthand notations:} We note that by chain rule, $\nabla_{w=w_t} \phi(X_t w,R_t)= X_t^{\top} \nabla_{s^{w_t}_t} \phi(s^{w_t}_t,R_t)$, where $s_t^{w_t}= X_t w_t$.  Since $X_t$ is deterministic in our setting, we focus on unbiased estimators of $\nabla_{s^{w_t}_t} \phi(s^{w_t}_t,R_t)$ and take a matrix-vector product with $X_t$. To reduce notational clutter in our derivations, we drop $w$ from $s^w$ and the subscript $t$ throughout. Thus, in our derivations, $\tilde{z}= \tilde{z}_t$, $X=X_t$, $s= s_t^{w_t}$ (and not $\tilde{s}_t$), $\sigma= \tilde{\sigma}_t$ (and not $\sigma_t$), $R= R_t$, $e_i$ is standard basis vector in $\mathbb{R}^m$ along coordinate $i$ and $p(\cdot)$ as in Eq.~\ref{eq:shortprob} with $\mathbb{P}= \mathbb{P}_t$ where $\mathbb{P}_t$ is the distribution in round $t$ in Alg.~\ref{alg:RTop-kF}.
%We would like to point out that for ranking surrogates of the form $\phi(s^{w_t}_t,R_t)$, where $s^{w_t}_t= X_t w_t$, by chain rule, we have $\nabla_{w=w_t} \phi(X_t w,R_t)= X_t^{\top} \nabla_{s^{w_t}_t} \phi(s^{w_t}_t,R_t)$.  Since $X_t$ is deterministic in our setting, we will construct unbiased estimators of $\nabla_{s^{w_t}_t} \phi(s^{w_t}_t,R_t)$ for various surrogates $\phi$. To reduce notational clutter, we drop $w$ from $s^w$ and the subscript $t$ throughout, since we are concerned with unbiased estimator within a single round of the online game. Thus, when we construct unbiased estimator of $\nabla_s \phi(s,R)$, we mean $\nabla_{s^{w_t}_t} \phi(s^{w_t}_t, R_t)$.

\subsection{ Convex Surrogates}
\label{convexsurrogates}
{\bf Pointwise Method:} We will construct the unbiased estimator of the gradient of squared loss \citep{cossock2006subset}: $\phi_{sq}(s,R)= \|s-R\|_2^2$. The gradient $\nabla_s \phi_{sq}(s,R)$ is $2(s-R) \in \mathbb{R}^m$.  As we have already demonstrated in the example following Lemma~\ref{unbiasedestimator}, {\bf we can construct unbiased estimator of $R$ from top-1 feedback }($\{\sigma,R_{\sigma(1)}\}$). Concretely, the unbiased estimator is:
\[
{\bf\tilde{z}}=  X^{\top} \left(2 \left(s  - \dfrac{R_{\sigma(1)} e_{\sigma(1)}}{p(\sigma(1))} \right) \right) .
\] 

{\bf Pairwise Method:} We will construct the unbiased estimator of the gradient of hinge-like surrogate in RankSVM \citep{joachims2002}: $\phi_{svm}(s,R)= \sum_{i \neq j =1} \mathbbm{1}(R_i>R_j) \max (0, 1 + s_j -s_i)$. The gradient is given by $\nabla_s \phi_{svm}(s,R)=\sum_{i \neq j =1}^m  \mathbbm{1}(R_i>R_j) \mathbbm{1}(1+s_j>s_i) (e_j- e_i) \in \mathbb{R}^m$. Since $s$ is a known quantity, from Lemma~\ref{unbiasedestimator}, we can construct $F(R)$ as follows: $F(R)= F_s(R)= \sum_{i \neq j =1}^m h_{s, i,j}(R_i,R_j)$, where $ h_{s, i,j}(R_i,R_j)= \mathbbm{1}(R_i >R_j) \mathbbm{1}(1+s_j>s_i) (e_j - e_i)$.  Since $F_s(R)$ is decomposable over 2 coordinates of $R$ at a time, {\bf we can construct an unbiased estimator from top-2 feedback} ($\{\sigma, R_{\sigma(1)}, R_{\sigma(2)}\}$). The unbiased estimator is:
\[
\small
\begin{split}
&{\bf \tilde{z}}=  \\
& X^{\top} \left(\dfrac{h_{s,\sigma(1),\sigma(2)}(R_{\sigma(1)}, R_{\sigma(2)}) + h_{s,\sigma(2),\sigma(1)}(R_{\sigma(2)}, R_{\sigma(1)})}{p(\sigma(1),\sigma(2))+ p(\sigma(2),\sigma(1))}\right) .
\end{split}
\]

We note that the unbiased estimator was constructed from top-2 feedback. The following lemma, in conjunction with the necessary condition of Lemma~\ref{unbiasedestimator} shows that it is the minimum information required to construct the unbiased estimator.

\begin{lemma}
\label{RankSVM}
The gradient of RankSVM surrogate, i.e., $\phi_{svm}(s,R)$ cannot be decomposed over 1 coordinate of R at a time. 
\end{lemma}
 
%${\bf \tilde{z}}$= $ X^{\top}\left(\dfrac{\mathbbm{1}(R_{\sigma(1)}>R_{t\sigma(2)})\mathbbm{1}(1+s_{\sigma(2)}>s_{\sigma(1)})(e_{\sigma(2)}- e_{\sigma(1)}) + \mathbbm{1}(R_{\sigma(2)}>R_{t\sigma(1)})\mathbbm{1}(1+s_{\sigma(1)}>s_{\sigma(2)})(e_{\sigma(1)}- e_{\sigma(2)})}{p(\sigma(1),\sigma(2))+ p(\sigma(2),\sigma(1))}\right)$ $\in \reals^m$ , where $p(\sigma(1), \sigma(2))= \sum\limits_{\pi \in S_m} \mathbb{P} (\pi) \mathbbm{1}(\pi(1)= \sigma(1), \pi(2)= \sigma(2))$ ($\mathbb{P}= \mathbb{P}_t$ (Step-5 in Alg.~\ref{alg:RTop-kF})).
% 
{\bf Listwise Method:}  Convex surrogates developed for listwise methods of learning to rank are defined over the entire score vector and relevance vector. Gradient of a surrogate cannot usually be decomposed over coordinates of the relevance vector. We will focus on the cross-entropy surrogate used in the highly cited ListNet \citep{cao2007learning} ranking algorithm and show how a very natural modification to the surrogate makes its gradient estimable in our partial feedback setting.

The authors of the ListNet method use a cross-entropy surrogate on two probability distributions on permutations, induced by score and relevance vector respectively. More formally, the surrogate is defined as follows\footnote{The ListNet paper actually defines a family of losses based on probability models for top $r$ documents, with $r \le m$. We use $r=1$ in our definition since that is the version implemented in their experimental results.}. Define $m$ maps from $\reals^m$ to $\reals$ as: $P_j(v) = \exp(v_j)/\sum_{j=1}^m \exp(v_j)$ for $j \in [m]$. Then, for score vector $s$ and relevance vector $R$, 
$\listnet(s,R) = - \sum_{i=1}^m P_i(R) \log P_i(s)$ and $\nabla_s \listnet(s,R) = \sum_{i=1}^m \left(- \frac{\exp(R_i)}{\sum_{j=1}^m \exp(R_j)} + \frac{\exp(s_i)}{\sum_{j=1}^m \exp(s_{j})} \right) e_i$. We have the following lemma about the gradient of $\phi_{LN}$.
\begin{lemma}
\label{listnet}
The gradient of ListNet surrogate $\phi_{LN}(s,R)$ cannot be decomposed over $k$, for $k = 1,2$, coordinates of R at a time.
\end{lemma}

In fact, an examination of the proof of the above lemma reveals that decomposability at any $k < m$ does not hold for the gradient of LisNet surrogate, though we only prove it for $k=1,2$ (since feedback for top $k$ items with $k>2$ does not seem practical).  Due to Lemma~\ref{unbiasedestimator}, this means that if we want to run Alg.~\ref{alg:RTop-kF} under top-$k$ feedback, a modification of ListNet is needed. We now make such a modification. 

We first note that the cross-entropy surrogate of ListNet can be easily obtained from a standard divergence, viz. Kullback-Liebler divergence. Let $p, q \in \reals^m$ be 2 probability distributions ($\sum_{i=1}^m p_i= \sum_{i=1}^m q_i= 1$). Then $KL(p,q)= \sum_{i=1}^m p_i \log(p_i) - \sum_{i=1}^m p_i \log(q_i) - \sum_{i=1}^m p_i + \sum_{i=1}^m q_i$. Taking $p_i= P_i(R)$ and $q_i= P_i(s)$, $\forall \ i\in [m]$ (where $P_i(v)$ is as defined in $\listnet$) and noting that $\listnet(s,R)$ needs to be minimized w.r.t. $s$ (thus we can ignore the $\sum_{i=1}^m p_i \log(p_i)$ term in $KL(p,q)$), we get the cross entropy surrogate from KL.

Our natural modification now easily follows by considering KL divergence for \emph{un-normalized} vectors (it should be noted that KL divergence is an instance of a Bregman divergence). Define $m$ maps from $\reals^m$ to $\reals$ as: $P'_j(v) = \exp(v_j)$ for $j \in [m]$. Now define $p_i= P'_i(R)$ and $q_i = P'_i(s)$. Then, the modified surrogate is $\phi_{KL}(s,R)$ is:
$$\sum\limits_{i=1}^m e^{R_i} \log(e^{R_i}) - \sum\limits_{i=1}^m e^{R_i} \log(e^{s_i}) - \sum\limits_{i=1}^m e^{R_i} + \sum\limits_{i=1}^m e^{s_i} ,$$
and $\sum\limits_{i=1}^m \left(\exp(s_i) - \exp(R_i) \right) e_i$ is its gradient w.r.t. $s$.
Note that $\phi_{KL}(s,R)$ is non-negative and convex in $s$. Equating gradient to ${\bf 0} \in \reals^m$, at the minimum point, $s_i= R_i, \ \forall \ i \in [m]$. Thus, the sorted order of optimal score vector agrees with sorted order of relevance vector and it is a valid ranking surrogate. 

Now, from Lemma~\ref{unbiasedestimator}, we can construct $F(R)$ as follows: $F(R)= F_s(R)= \sum_{i =1}^m h_{s, i}(R_i)$, where $h_{s, i}(R_i)= \left(\exp(s_i) - \exp(R_i) \right) e_i$. Since $F_s(R)$ is decomposable over 1 coordinate of $R$ at a time, {\bf we can construct an unbiased estimator from top-1 feedback} ($\{\sigma, R_{\sigma(1)} \}$). The unbiased estimator is:
\[
{\bf \tilde{z}}= X^{\top} \left( \dfrac{(\exp(s_{\sigma(1)}) - \exp(R_{\sigma(1)}))e_{\sigma(1)}}{p(\sigma(1))}\right)
\]
 
{\bf Other Listwise Methods:} As we mentioned before, most listwise convex surrogates will not be suitable for Alg.~\ref{alg:RTop-kF} with top-k feedback. For example, the class of popular listwise surrogates that are developed from structured prediction perspective \citep{chapelle2007, yue2007} cannot have unbiased estimator of gradients from top-k feedback since they are based on maps from full relevance vectors to full rankings and thus cannot be decomposed over $k=1$ or $2$ coordinates of $R$. It does not appear they have any natural modification to make them amenable to our approach.

\subsubsection{Non-convex Surrogate}
\label{nonconvexsurrogate}
We provide an example of a non-convex surrogate for which Alg.~\ref{alg:RTop-kF} is applicable (however it will not have any regret guarantees due to non-convexity). We choose the SmoothDCG surrogate given in \citep{chapelle2010gradient}, which has been shown to have very competitive empirical performance. SmoothDCG, like ListNet, defines a family of surrogates, based on the cut-off point of DCG (see original paper \citep{chapelle2010gradient} for details). We consider  SmoothDCG@1, which is the smooth version of DCG@1 (i.e., DCG which focuses just on the top-ranked document). 
The surrogate is defined as: $\phi_{SD}(s,R)= \frac{1}{\sum_{j=1}^m \exp(s_j/{\epsilon})} \sum_{i=1}^m G(R_i) \exp(s_i/{\epsilon})$, where $\epsilon$ is a (known) smoothing parameter and $G(a)= 2^a-1$. The gradient of the surrogate is:
%\begin{equation*}
%\nabla_s \phi_{sdcg}(s,R)= \frac{1}{\epsilon \sum_{j'=1}^m \exp(s_{j'}/{\epsilon})}\sum_{i=1}^m G(R_i) \left( \sum_{j=1}^m [ \exp(s_i/{\epsilon}] e_j \right)
%\end{equation*}
\[
\small
\begin{split}
& [\nabla_s \sdcg(s,R)] = \sum_{i=1}^m h_{s,i}(R_i), \   h_{s,i}(R_i)=  \\  
& G(R_i) \left(\sum_{j=1}^m [ \frac{1}{\epsilon} \frac{\exp(s_i/\epsilon)}{\sum_{j} \exp(s_{j'}/\epsilon)} \mathbbm{1}_{(i = j)} 
- \frac{1}{\epsilon}\frac{\exp((s_i+s_j)/\epsilon)}{(\sum_{j'} \exp(s_{j'}/\epsilon))^2} ]e_j \right)
\end{split}
\]
Using Lemma~\ref{unbiasedestimator}, we can write $F(R)= F_s(R)= \sum_{i=1}^m h_{s,i}(R_i)$ where $h_{s,i}(R_i)$ is defined above. Since $F_s(R)$ is decomposable over 1 coordinate of $R$ at a time, {\bf we can construct an unbiased estimator from top-1 feedback} ($\{\sigma, R_{\sigma(1)} \}$), with unbiased estimator being:
\[
\small
\begin{split}
& {\bf \tilde{z}} =  X^{\top} \left(\dfrac{G(R_{\sigma(1)})}{p(\sigma(1))} (*) \right)\\
& (*)= \sum_{j=1}^m [ \dfrac{1}{\epsilon} \dfrac{\exp(s_{\sigma(1)}/\epsilon)}{\sum_{j'} \exp(s_{j'}/\epsilon)} \mathbbm{1}_{(\sigma(1) = j)}
- \frac{1}{\epsilon}\dfrac{\exp((s_{\sigma(1)}+s_j)/\epsilon)}{(\sum_{j'} \exp(s_{j'}/\epsilon))^2} ]e_j 
\end{split}
\]

\subsection {Computational Complexity of Algorithm~\ref{alg:RTop-kF}}
Three of the four key steps governing the complexity of Alg.~\ref{alg:RTop-kF}, i.e., construction of $\tilde{s}_t$, $\tilde{\sigma}_t$ and sorting can all be done in $O(m \log(m))$ time. Construction of estimator is even simpler. The only bottleneck could have been calculations of $p(\sigma(1))$ in squared loss, (modified) ListNet loss and SmoothDCG loss, and $p(\sigma(1),\sigma(2))$ in RankSVM loss, since they involve sum over permutations. However, they have a compact representation, i.e., $p(\sigma(1))= 1 -\gamma + \frac{\gamma}{m}$ and $p(\sigma(1),\sigma(2))= 1 -\gamma +\frac{2 \gamma}{m(m-1)}$. The calculations follow easily due to the nature of $\mathbb{P}_t$ (step-6 in algorithm) which put equal weights on all permutations other than $\sigma_t$.

\subsection{Regret Bounds}
The underlying deterministic part of our algorithm is online gradient descent (OGD) \citep{zinkevich2003online}. The regret of OGD, run with unbiased estimator of gradient of a {\bf convex} function, as given in Theorem 3.1 of \citep{flaxman2005}, in our problem setting is:
\begin{equation}
\small
\label{eq:flaxman}
\begin{split}
\E \left[\sum_ {t=1}^T \phi(X_tw_t, R_t) \right] \le & \underset{w:\|w\|_2 \le U} {\min} \sum_{t=1}^T \phi(X_t w,R_t) + \\
& \frac{U^2}{2\eta} +  \frac{\eta}{2} \E \left[ {\sum_{t=1}^T \|\tilde{z}_t\|_2^2} \right]
\end{split} 
\end{equation}
where $\tilde{z}_t$ is unbiased estimator of $\nabla_{w=w_t}\phi(X_t w,R_t)$,  conditioned on past events, $\eta$ is the learning rate and the expectation is taken over all randomness in the algorithm.

However, from the perspective of the loss $\phi(\tilde{s}_t,R_t)$ incurred by Alg.~\ref{alg:RTop-kF}, at each round $t$, the RHS above is not a valid upper bound. The algorithms plays the score vector suggested by OGD ($\tilde{s}_t= X_t w_t$) with probability $1-\gamma$ (exploitation) and plays a randomly selected score vector (i.e., a draw from the uniform distribution on $[0,1]^m$), with probability $\gamma$ (exploration). Thus, the expected number of rounds in which the algorithm does not follow the score suggested by OGD is $\gamma T$, leading to an extra regret\footnote{The instantaneous loss suffered at each of the exploration round can be maximum of $O(1)$, as long as $\phi(s,R)$ is bounded, $\forall \ s$ and $\forall \ R$. This is true because the score space is $\ell_2$ norm bounded, maximum relevance grade is finite in practice and we consider Lipschitz, convex surrogates.} of order $\gamma T$. Thus, we have
\begin{equation}
\label{eq:exploration}
\E \left[\sum_ {t=1}^T \phi(\tilde{s}_t, R_t) \right] \le \E \left[\sum_ {t=1}^T \phi(X_t w_t, R_t) \right]  + O \left(\gamma T \right) 
\end{equation}

We first control $\E_t \|\tilde{z}_t\|_2^2$, for all convex surrogates considered in our problem (we remind that $\tilde{z}_t$ is the estimator of a gradient of a surrogate, calculated at time $t$. In Sec~\ref{convexsurrogates} , we omitted showing $w$ in $s^w$ and index $t$).
To get bound on  $\E_t \|\tilde{z}_t\|_2^2$, we used the following norm relation that holds for any matrix $X$ \citep{bhaskara2011}: $\|X\|_{p \to q}= \underset{v \neq 0}{\sup} \frac{\|Xv\|_q}{\|v\|_p}$, where $q$ is the dual exponent of $p$ (i.e., $\tfrac{1}{q}+\tfrac{1}{p} = 1$), and the following lemma derived from it:
\begin{lemma}
\label{normexpr}
For any $1 \leq p \leq \infty$, $\| X^{\top} \|_{1 \to p} = \| X \|_{q \to \infty} = \max_{j=1}^m \| X_{j:} \|_p$, 
%\begin{equation*}
%\| X^{\top} \|_{1 \to p} = \| X \|_{q \to \infty} = \max_{j=1}^m \| X_{j:} \|_p \ ,
%\end{equation*}
where $X_{j:}$ denotes $j$th row of $X$ and $m$ is the number of rows of matrix.
\end{lemma}

\begin{lemma}
\label{expectednorm}
For parameter $\gamma$ in Alg.~\ref{alg:RTop-kF} , $R_D$ being the bound on $\ell_2$ norm of the feature vectors (rows of document matrix $X$), $m$ being the upper bound on number of documents per query, $U$ being the radius of the Euclidean ball denoting the space of ranking parameters and $R_{\max}$ being the maximum possible relevance value (in practice always $\le$ 5), let $C^{\phi} \in \{C^{sq}, C^{svm}, C^{KL}\}$ be polynomial functions of $R_D, m, U, R_{max}$, where the degrees of the polynomials depend on the surrogate ($\phi_{sq}, \phi_{svm}, \phi_{KL}$), with no degree ever greater than four. Then we have, 
\begin{equation}
\label{eq:expectednorm}
\begin{split}
 \E_t \left[\|\tilde{z}_t\|_2^2 \right] \le \dfrac{C^{\phi}}{\gamma}
\end{split}
\end{equation}
\end{lemma}

Plugging Eq.~\ref{eq:expectednorm} and Eq.~\ref{eq:exploration} in Eq.~\ref{eq:flaxman}, and optimizing over $\eta$ and $\gamma$, (which gives $\eta= O(T^{-2/3})$ and $\gamma= O(T^{-1/3})$), we get the final regret bound.

\begin{theorem}
\label{theoryboundinpartial}
For any sequence of instances and labels $(X_t,R_t)_{\{t \in [T]\}}$, applying Alg.~\ref{alg:RTop-kF} with top-1 feedback for $\phi_{sq}$ and $\phi_{KL}$ and top-2 feedback for $\phi_{svm}$,  will produce the following bound on the regret for any of the three surrogates: 
\begin{equation}
\small
\begin{split}
\E \left[\sum_{t=1}^T \phi(\tilde{s}_t, R_t) \right] -  \underset{w:\|w\|_2 \le U}{\min} \sum_{t=1}^T \phi(X_t w,R_t)  \le C^{\phi} O \left(T^{2/3} \right)
\end{split}
\end{equation}
where $C^{\phi}$ is a surrogate dependent function, as described in Lemma~\ref{expectednorm} , and  expectation is taken over underlying randomness of the algorithm, over $T$ rounds.
\end{theorem}

{\bf Discussion:} It is known that online bandit games are special instances of partial monitoring games. For bandit online convex optimization problems with Lipschitz, convex surrogates, the best regret rate known so far, that can be achieved by an efficient algorithm, is $O(T^{3/4})$ (however, see the work of \cite{bubeck2015multi} for a non-constructive $O(\log^4(T) \sqrt{T})$ bound). Surprisingly, Alg.~\ref{alg:RTop-kF}, when applied in a partial monitoring setting to the Lipschitz, convex surrogates that we have listed, achieves a better regret rate than what is known in the bandit setting. Moreover, as we show subsequently, for an entire class of Lipschitz convex surrogates (subclass of NDCG calibrated surrogates), sub-linear (in $T$) regret is not even achievable. Thus, our work indicates that even within the class of Lipschitz, convex surrogates, regret rate achievable is dependent on the structure of surrogates; something that does not arise in bandit convex optimization. 

\section{Impossibility of Sublinear Regret  for NDCG Calibrated Surrogates}
Learning to rank methods optimize surrogates to learn a ranking function, even though performance is measured by target measures like NDCG. This is done because direct optimization of the measures lead to NP-hard optimization problems. One of the most desirable properties of any surrogate is \emph{calibration}, i.e., the surrogate should be calibrated w.r.t the target \citep{bartlett2006convexity}. Intuitively, it means that a function with small expected surrogate loss on unseen data should have small expect target loss on unseen data.  We focus on NDCG calibrated surrogates (both convex and non-convex) that have been characterized by \cite{ravikumar2011ndcg}. We first state the necessary and sufficient condition for a surrogate to be calibrated w.r.t NDCG. For any score vector $s$ and distribution $\eta$ on relevance space $\mathcal{Y}$, let $\bar{\phi}(s,\eta)= \E_{R \sim \eta} \phi(s,R)$. Moreover, we define $G({\bf R})= (G(R_1),\ldots,G(R_m))^{\top}$. 
\begin{theorem}
\label{calibration}
\cite[Thm. 6]{ravikumar2011ndcg} A surrogate $\phi$ is NDCG calibrated iff for any distribution $\eta$ on relevance space $\mathcal{Y}$, there exists an invertible, order preserving map $g: \reals^m \mapsto \reals^m$ s.t. the unique minimizer $s^{*}_{\phi}(\eta)$ can be written as
\begin{equation}
\label{eq:calibration}
s^*_{\phi}(\eta)= g \left( \E_{R \sim \eta} \left[\frac{G({\bf R})}{Z_m(R)}\right] \right) .
\end{equation}
\end{theorem}
Informally, Eq.~\ref{eq:calibration} states that $\argsort(s^*_{\phi}(\eta)) \subseteq \argsort(\E_{R \sim \eta} \left[\tfrac{G({\bf R})}{Z_m(R)}\right])$ 
\cite{ravikumar2011ndcg} give concrete examples of NDCG calibrated surrogates, including how some of the popular surrogates can be converted into NDCG calibrated ones: e.g., the NDCG calibrated version of squared loss is $\|s- \frac{G({\bf R})}{Z_m(R)}\|_2^2$.  

We now state the \emph{impossibility} result for the class of NDCG calibrated surrogates with top-1 feedback.

\begin{theorem}
\label{impossiblegame}
Fix the online learning to rank game with top-1 feedback and any NDCG calibrated surrogate. Then, for every learner's algorithm, there exists an adversary strategy s.t. the learner's expected regret is $\Omega(T)$. 
\end{theorem}

Note that our result is for top-1 feedback. Minimax regret for the problem setting with top-$k$ feedback, with $k \ge 2$ remains an open question.

\begin{proof} ({\it Sketch})
The proof builds on the proof of hopeless finite action partial monitoring games given by \cite{piccolboni2001discrete}. An examination of their proof of Thm.\ 3 indicates that for hopeless games, there have to exist two probability distributions (over adversary's actions), which are indistinguishable in terms of feedback but the optimal learner's actions for the distributions are different. We first provide a mathematical explanation as to why such existence lead to hopeless games. Then, we provide a characterization of indistinguishable probability distributions in our problem setting, and then exploit the characterization of optimal actions for NDCG calibrated surrogates (Thm.~\ref{calibration}) to explicitly construct two such probability distributions. This proves the result.
\end{proof}
 
We note that the proof of Thm.\ 3 of \cite{piccolboni2001discrete} cannot be directly extended to prove the impossibility result because it relies on constructing a connected graph on vertices defined by neighboring actions of learner. In our case, due to the continuous nature of learner's actions, the graph will be an empty graph and proof will break down.

\section{Empirical Results}
\begin{figure}[t]
\minipage{0.53\textwidth}
  \includegraphics[height=48mm, width=\linewidth]{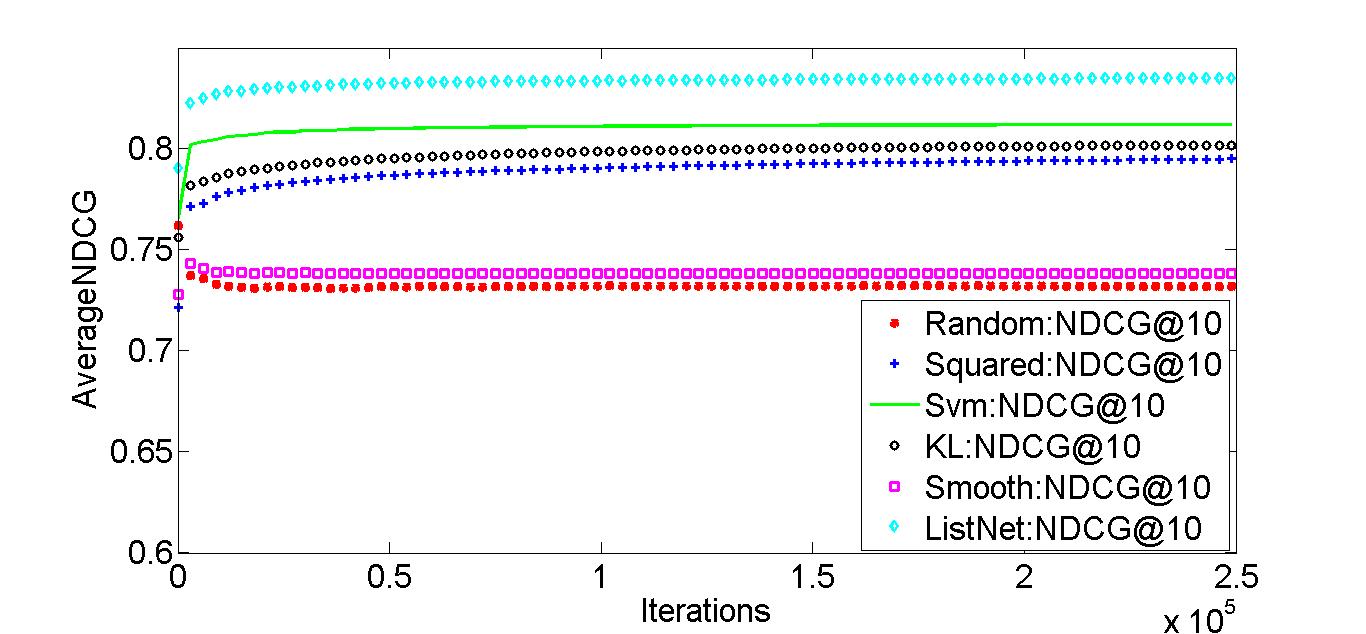}
 \endminipage\hfill
\minipage{0.53\textwidth}
  \includegraphics[height=48mm, width=\linewidth]{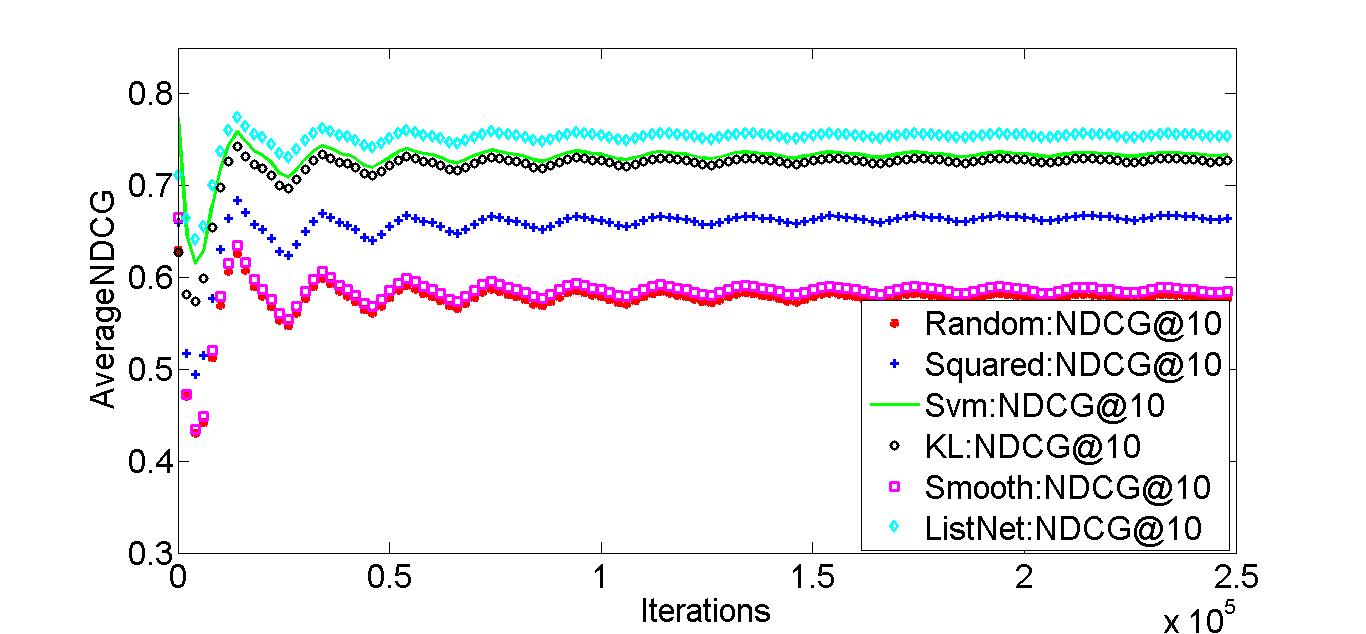}
 \endminipage\hfill
\caption{Average $\text{NDCG}$@$10$  values for different algorithms, for {\bf Yandex (top)} and {\bf Yahoo (bottom)}. ListNet:NDCG$@$10 (in cyan) is a {\bf full feedback} algorithm and Random:NDCG$@$10 (in red) is {\bf no feedback} algorithm.}\label{Fig1}
\end{figure}

{\bf Objective:} We conducted experiments on benchmark datasets to demonstrate the performance of ranking functions that are learnt  from partial feedback. As stated before, though Alg.~\ref{alg:RTop-kF} is designed to minimize surrogate based regret, the users only care about the ranking presented to them, and indeed the algorithm interacts with users only through ranked lists. We tested the quality of the ranked lists, and hence the performance of the evolving ranking functions, against the full relevance vectors via $\text{NDCG}_{10}$.\\
{\bf Ranking functions compared:} We applied Alg.~\ref{alg:RTop-kF}, with top-1 feedback, on Squared, KL and SmoothDCG surrogates, and with top-2 feedback, on the RankSVM surrogate. \emph{Since our work is based on a novel feedback model, the performance of Alg.\ \ref{alg:RTop-kF} could not be directly compared with any published baseline}. So, based on the objective of our work, we selected two different ranking methods for comparison.  The first one is the online version ListNet ranking algorithm, with full relevance vector revealed at end of every round. ListNet is not only one of the most cited ranking algorithms (over 700 citations according to Google Scholar), but also one of the most validated algorithms \citep{tax2015}. We emphasize that some of the ranking algorithms, which have shown better empirical performance than ListNet, are usually based on non-convex surrogates with complex ranking functions. These algorithms cannot usually be converted into online algorithms which learn from streaming data. The second one is a fully random algorithm which outputs a uniformly at random ranking of documents at each round. \emph{Effectively, we are comparing Alg~\ref{alg:RTop-kF}, which learns from highly restricted feedback, with an algorithm which learns from full feedback and another algorithm which receives no feedback.}  \\
{\bf Datasets:} We compared the various ranking functions on two large scale commercial datasets. They were Yahoo's Learning to Rank Challenge dataset \citep{chapelle2011yahoo} and a dataset published by Russian search engine Yandex \citep{Yandex}. The Yahoo dataset has 19944 unique queries with 5 distinct relevance levels, while Yandex has 9126 unique queries with 5 distinct relevance levels.\\
{\bf Setting of experiments}: We selected time horizon $T=$ 250,000 iterations for our experiments (thus, each algorithm went over each dataset multiple times). All the online algorithms, other than the fully random one, involve learning rate $\eta$ and exploration parameter $\gamma$ (Full information ListNet does not involve $\gamma$ and SmoothDCG has an additional smoothing parameter $\epsilon$). While obtaining our regret guarantees, we had established that $\eta=O(T^{-2/3})$ and $\gamma= O(T^{-1/3})$ and thus, in our experiments, for each instance of Alg.~\ref{alg:RTop-kF}, we selected $\eta=\frac{1}{T^{2/3}}$ and $\gamma=\frac{1}{T^{1/3}}$. We fixed $\epsilon=0.01$. For ListNet, we selected $\eta=\frac{1}{T^{1/2}}$, since regret guaratnee in OGD is established with $\eta=O(T^{-1/2})$. We plotted average $\text{NDCG}_{10}$ against time, where average  $\text{NDCG}_{10}$ at time $t$ is the cumulative $\text{NDCG}_{10}$ up to time $t$, divided by $t$.\\ 
{\bf Observations:} In both the datasets, ListNet, with full information, has highest average NDCG value throughout. However, Alg.~\ref{alg:RTop-kF}, with the convex surrogates, produce competitive performance. In fact, in the Yahoo dataset, our algorithms, with RankSVM and KL, are very close to the performance of ListNet. RanSVM does better than the other surrogates, since the estimator of RankSVM gradient is constructed from top-2 feedback, leading to lower variance. KL, being listwise in nature, does better than squared loss. \emph{Crucially, our algorithms, based on all three convex surrogates, perform significantly better than the purely random algorithm, and are much closer to full feedback ListNet in performance, despite being much closer to the purely random algorithm in terms of feedback}. Our algorithm, with SmoothDCG, on the other hand, produce poor performance. We believe the reason is the non-convexity of the surrogate, which leads to the optimization procedure possibly getting stuck at a local minima. In batch setting, such problem is avoided by an annealing technique that successively reduces $\epsilon$. We are not aware of an analogue in an online setting. Possible algorithms optimizing non-convex surrogates in an online manner, which require gradient of the surrogate, may be adapted to this partial feedback setting. 

%In both the datasets, RankSVM has uniformly highest $NDCG_{10}$ value, followed by KL surrogate, followed by squared loss, all of which perform better than the baseline (Random). SmoothDCG performs slightly better than random. We first focus on the performance of the convex surrogates and compare them to the baseline. Since RankSVM is constructed from top-2 feedback, the unbiased estimator will have lower variance, than the estimators constructed from top-1 feedback. Thus, RankSVM's empirical performance is expected. KL, being listwise, does better than squared loss, which is pointwise, as expected. All three convex surrogates does better than the baseline, with a considerable, statistically significant, gain.  We believe the reason SmoothDCG is performing poorly is because of non-convexity of the surrogate, which leads to the optimization procedure possibly getting stuck at a local minima. In batch setting, such problem is avoided by the annealing technique. We are not aware of an analogue in an online setting. \emph{Our main intention for including the non-convex surrogate was to demonstrate the construction of unbiased estimator of gradient of a non-convex surrogate based on limited feedback}. Possible algorithms optimizing non-convex surrogates in an online manner, which require gradient of the surrogate, may be adapted to this partial feedback setting. 

% If paper is accepted, acknowledge NSF.

\bibliography{RankingBib}
\bibliographystyle{plainnat}

\onecolumn % Switch to one column format
\newpage

\section{Supplementary}

{\bf Proof of Lemma~\ref{unbiasedestimator}}: We restate the lemma before giving the proof, for ease of reading:

{\bf Lemma 1}: Let $F: \mathbb{R}^m \mapsto \mathbb{R}^a$ be a vector valued function, where $m\ge 1$, $a\ge 1$. For a fixed $x \in \mathbb{R}^m$, let $k$ entries of $x$ be observed at random. That is, for a fixed probability distribution $\mathbb{P}$ and some random $\sigma \sim \mathbb{P}(S_m)$, observed tuple is $\{\sigma, x_{\sigma(1)}, \ldots, x_{\sigma(k)}\}$. The necessary condition for existence of an unbiased estimator of $F(x)$, that can be constructed from $\{\sigma, x_{\sigma(1)}, \ldots, x_{\sigma(k)}\}$, is that it should be  possible to decompose $F(x)$ over $k$ (or less) coordinates of $x$ at a time. That is, $F(x)$ should have the following structure:
\begin{equation*}
F(x)= \sum\limits_{(i_1,i_2,\ldots,i_{\ell}) \in \ \perm{m}{\ell}} h_{i_1,i_2,\ldots,i_{\ell}}(x_{i_1}, x_{i_2},\ldots, x_{i_{\ell}}) 
\end{equation*}
where $\ell \le k$, $\perm{m}{\ell}$ is $\ell$ permutations of $m$ and $h: \mathbb{R}^{\ell} \mapsto \mathbb{R}^a$.
Moreover, when $F(x)$ can be written in form of Eq~\ref{eq:decoupling} , with $\ell=k$, an unbiased estimator of $F(x)$, based on $\{\sigma, x_{\sigma(1)}, \ldots, x_{\sigma(k)}\}$, is, 
%\begin{equation}
%\label{eq:unbiasedestimator}
%g(\sigma, x_{\sigma(1)}, \ldots, x_{\sigma(k)})=  \dfrac{h_{\sigma(1),\ldots, \sigma(k)}(x_{\sigma(1)}, \ldots, x_{\sigma(k)})}{p(\sigma(1),\ldots,\sigma(k))} 
%\end{equation}
%where $p(\sigma(1),\ldots, \sigma(k)) = \sum\limits_{\pi \in S_m}\mathbb{P}(\pi) \mathbbm{1}(\pi(1)= \sigma(1),\ldots,\pi(k)=\sigma(k))$ (as in Eq.~\ref{eq:shortprob} )
\begin{equation*}
g(\sigma, x_{\sigma(1)}, \ldots, x_{\sigma(k)})=  \dfrac{\sum\limits_{(j_1,j_2,\ldots, j_k) \in S_k}h_{\sigma(j_1),\ldots, \sigma(j_k)}(x_{\sigma(j_1)}, \ldots, x_{\sigma(j_k)})}{\sum\limits_{\substack{(j_1,\ldots, j_k) \in S_k}} p(\sigma(j_1),\ldots,\sigma(j_k))} 
\end{equation*}
where $S_k$ is the set of $k!$ permutations of $[$k$]$ and $p(\sigma(1),\ldots, \sigma(k))$ is as in Eq~\ref{eq:shortprob} .

\begin{proof}
For a fixed $x \in \mathbb{R}^m$ and probability distribution $\mathbb{P}$, let the random permutation be $\sigma \sim \mathbb{P}(S_m)$ and the observed tuple be $\{\sigma, x_{\sigma(1)}, \ldots, x_{\sigma(k)}\}$. Let $\hat{G}= G(\sigma, x_{\sigma(1)}, \ldots, x_{\sigma(k)})$ be an unbiased estimator of $F(x)$ based on the random observed tuple. Taking expectation, we get:
\begin{equation*}
\begin{aligned}
\begin{split}
F(x)= \E_{\sigma \sim \mathbb{P}} \left[\hat{G} \right]& = \sum_{\pi \in S_m} \mathbb{P}(\pi) G(\pi, x_{\pi(1)}, \ldots, x_{\pi(k)}) \\
&= \sum_{(i_1, i_2, \ldots, i_k) \in \ \perm{m}{k}} \sum_{\pi \in S_m} \mathbb{P}(\pi) \mathbbm{1}(\pi(1)=i_1, \pi(2)=i_2, \ldots, \pi(k)= i_k) G(\pi, x_{i_1}, x_{i_2}, \ldots, x_{i_k}) 
\end{split} 
\end{aligned}
\end{equation*}
We note that $\mathbb{P}(\pi) \in [0,1]$ is independent of $x$ for all $\pi \in S_m$. Then we can use the following construction of function $h(\cdot)$:
\begin{equation*}
h_{i_1,i_2,\ldots, i_k} (x_{i_1}, \ldots, x_{i_k})=  \sum_{\pi \in S_m} \mathbb{P}(\pi) \mathbbm{1}(\pi(1)=i_1, \pi(2)=i_2, \ldots, \pi(k)= i_k) G(\pi, x_{i_1}, x_{i_2}, \ldots, x_{i_k}) 
\end{equation*}
 and thus,
\begin{equation*}
F(x)= \sum\limits_{(i_1,i_2,\ldots,i_{k}) \in \ \perm{m}{k}} h_{i_1,i_2,\ldots,i_{k}}(x_{i_1}, x_{i_2},\ldots, x_{i_{}}) 
\end{equation*}
Hence, we conclude that for existence of an unbiased estimator based on the random observed tuple, it should be  possible to decompose $F(x)$ over $k$ (or less) coordinates of $x$ at a time. The ``less than $k$'' coordinates arguement follows simply by noting that if $F(x)$ can be decomposed over $\ell$ coordinates at a time ($\ell <k$) and observation tuple is \{$\sigma, x_{\sigma(1)}, \ldots, x_{\sigma(k)})$\}, then any $k - \ell$ observations can be thrown away and the rest used for construction of the unbiased estimator.

The construction of the unbiased estimator proceeds as follows:

Let $F(x)= \sum_{i=1}^m h_i(x_i)$ and feedback is for top-1 item ($k=1$). The  unbiased estimator according to Lemma.~\ref{unbiasedestimator} is:

\begin{equation*}
g(\sigma,x_{\sigma(1)})= \dfrac{h_{\sigma(1)}(x_{\sigma(1)})}{p(\sigma(1))} = \dfrac{h_{\sigma(1)}(x_{\sigma(1)})}{\sum_{\pi} \mathbb{P}(\pi) \mathbbm{1}(\pi (1)=\sigma(1))}
\end{equation*}

Taking expectation w.r.t. $\sigma$, we get:
\begin{equation*}
\E_{\sigma}[g(\sigma,x_{\sigma(1)})]= \sum_{i=1}^m \dfrac{h_{i}(x_{i}) (\sum_{\pi} \mathbb{P}(\pi) \mathbbm{1}(\pi(1)=i))}{\sum_{\pi} \mathbb{P}(\pi) \mathbbm{1}(\pi(1)=i)}= \sum_{i=1}^m h_{i}(x_i)= F(x)
\end{equation*}

Now, let $F(x)= \sum\limits_{i \neq j =1}^m h_{i,j}(x_{i},x_{j})$ and the feedback is for top-2 item ($k=2$). The  unbiased estimator according to Lemma.~\ref{unbiasedestimator} is:
\begin{equation*}
\begin{split}
g(\sigma,x_{\sigma(1)}, x_{\sigma(2)})& = \dfrac{h_{\sigma(1), \sigma(2)}(x_{\sigma(1)}, x_{\sigma(2)}) + h_{\sigma(2),\sigma(1)}(x_{\sigma(2)}, x_{\sigma(1)})}{p(\sigma(1), \sigma(2))+ p(\sigma(2), \sigma(1))}\\
% &= \dfrac{h_{\sigma(1), \sigma(2)}(x_{\sigma(1)}, x_{\sigma(2)}) + h_{\sigma(2),\sigma(1)}(x_{\sigma(2)}, x_{\sigma(1)})}{\sum_{\pi} \mathbb{P}(\pi) \mathbbm{1}(\pi (1)=\sigma(1), \pi(2)=\sigma(2)) + \sum_{\pi} \mathbb{P}(\pi) \mathbbm{1}(\pi (1)=\sigma(2), \pi(2)=\sigma(1))}
\end{split}
\end{equation*}

We will use the fact that for any 2 permutations $\sigma_1, \sigma_2$, which places the same 2 objects in top-2 positions but in opposite order, estimators based on $\sigma_1$ (i.e,  $g(\sigma_1, x_{\sigma_1(1)}, x_{\sigma_1(2)})$) and $\sigma_2$ (i.e, $g(\sigma_2, x_{\sigma_2(1)}, x_{\sigma_2(2)})$) have same numerator and denominator.  For eg., let $\sigma_1(1)=i, \sigma_1(2)=j$. Numerator and denominator for $g(\sigma_1, x_{\sigma_1(1)}, x_{\sigma_1(2)})$  are $h_{i,j}(x_i,x_j) + h_{j,i}(x_j,x_i)$ and $p(i,j)+ p(j,i)$ respectively. Now let $\sigma_2(1)=j, \sigma_2(2)=i$. Then numerator and denominator for $g(\sigma_2, x_{\sigma_2(1)}, x_{\sigma_2(2)})$ are $ h_{j,i}(x_j,x_i)+ h_{i,j}(x_i,x_j)$  and $p(j,i)+ p(i,j)$ respectively.

Then, taking expectation w.r.t. $\sigma$, we get:
\begin{equation*}
\begin{split}
\E_{\sigma}{g(\sigma,x_{\sigma(1)}, x_{\sigma(2)})} &= \sum_{i \neq j =1}^m \dfrac{(h_{i,j}(x_i,x_j) + h_{j,i}(x_j,x_i))p(i,j)}{p(i,j)+p(j,i)} \\
& = \sum_{i> j =1}^m \dfrac{(h_{i,j}(x_i,x_j) + h_{j,i}(x_j,x_i)) (p(i,j)+p(j,i))}{p(i,j)+p(j,i)} \\
&= \sum_{i>j=1}^m (h_{i,j}(x_i,x_j) + h_{j,i}(x_j,x_i)) = \sum_{i \neq j =1}^m h_{i,j}(x_i,x_j)  = F(x)
\end{split}
\end{equation*}

This chain of logic can be extended for any $k \ge 3$. Explicitly, for general $k \le m$, let $\mathbb{S}(i_1,i_2,\ldots, i_k)$ denote all permutations of the set $\{i_1, \ldots, i_k \}$. Then, taking expectation of the unbiased estimator will give:

\begin{equation*}
\begin{split}
&\E_{\sigma}{g(\sigma,x_{\sigma(1)}, \ldots, x_{\sigma(k)})} \\
& = \sum_{(i_1,i_2,\ldots,i_k) \in \ \perm{m}{k}} \dfrac{\left(\sum\limits_{(j_1,\ldots,j_k) \in \mathbb{S}(i_1,\ldots,i_k)}h_{j_1,\ldots,j_k}(x_{j_1},\ldots,x_{j_k})\right)p(i_1,\ldots,i_k)}{\sum\limits_{(j_1,\ldots,j_k) \in \mathbb{S}(i_1,\ldots,i_k)} p(j_1,\ldots,j_k)} \\
& = \sum_{i_1 > i_2 > \ldots > i_k =1}^m \dfrac{\left(\sum\limits_{(j_1,\ldots,j_k) \in \mathbb{S}(i_1,\ldots,i_k)}h_{j_1,\ldots,j_k}(x_{j_1},\ldots,x_{j_k})\right)\left(\sum\limits_{(j_1,\ldots,j_k) \in \mathbb{S}(i_1,\ldots,i_k)} p(j_1,\ldots,j_k)\right)}{\sum\limits_{(j_1,\ldots,j_k) \in \mathbb{S}(i_1,\ldots,i_k)} p(j_1,\ldots,j_k)} \\
&= \sum_{i_1 > i_2 > \ldots > i_k =1}^m \left(\sum\limits_{(j_1,\ldots,j_k) \in \mathbb{S}(i_1,\ldots,i_k)}h_{j_1,\ldots,j_k}(x_{j_1},\ldots,x_{j_k}) \right) = \sum\limits_{(i_1, i_2, \ldots, i_{k}) \in \ \perm{m}{k}} h_{i_1,i_2,\ldots,i_{k}}(x_{i_1}, x_{i_2},\ldots, x_{i_k})  = F(x)
\end{split}
\end{equation*}

{\bf Note}: For $k=m$, i.e., when the full feedback is received, the unbiased estimator is:

\begin{equation*}
\begin{split}
g(\sigma, x_{\sigma(1)}, \ldots, x_{\sigma(m)}) & =  \dfrac{\sum\limits_{(j_1,j_2,\ldots, j_m) \in S_m}h_{\sigma(j_1),\ldots, \sigma(j_m)}(x_{\sigma(j_1)}, \ldots, x_{\sigma(j_m)})}{\sum\limits_{\substack{(j_1,\ldots, j_m) \in S_m}} p(\sigma(j_1),\ldots,\sigma(j_m))} \\
&=  \dfrac{\sum\limits_{(i_1, i_2, \ldots, i_m) \in \ \perm{m}{m}} h_{i_1,\ldots, i_m}(x_{i_1}, \ldots, x_{i_m})}{1}= F(x)
\end{split}
\end{equation*}

Hence, with full information, the unbiased estimator of $F(x)$ is actually $F(x)$ itself, which is consistent with the theory of unbiased estimator.

\end{proof}

{\bf Proof of Lemma~\ref{normexpr}}: 

\begin{proof}
The first equality is true because
\begin{align*}
\| X^\top \|_{1 \to p} &= \sup_{v \neq 0} \frac{ \| X^\top v \|_p }{\| v \|_1} 
= \sup_{v \neq 0} \sup_{u \neq 0} \frac{ \inner{X^\top v, u} }{\| v \|_1 \|u\|_q} \\
&= \sup_{u \neq 0} \sup_{v \neq 0} \frac{ \inner{v, X u} }{\| v \|_1 \|u\|_q} 
= \sup_{u \neq 0} \frac{ \| X u\|_\infty }{\| u \|_q} 
= \| X \|_{q \to \infty} .
\end{align*}
The second is true because
\begin{align*}
\| X \|_{q \to \infty} &= \sup_{u \neq 0} \frac{ \| X u \|_\infty }{\|u\|_q }
= \sup_{u \neq 0} \max_{j=1}^m \frac{ |\inner{X_{j:},u}| }{\|u\|_q} \\
&= \max_{j=1}^m \sup_{u \neq 0} \frac{ |\inner{X_{j:},u}| }{\|u\|_q}
= \max_{j=1}^m \| X_{j:} \|_p . 
\end{align*} 
\end{proof}

{\bf Proof of Lemma~\ref{expectednorm}} : We restate the lemma before giving the proof: 

{\bf Lemma 5}:
For parameter $\gamma$ in Algorithm~\ref{alg:RTop-kF} , $R_D$ being the bound on $\ell_2$ norm of the feature vectors (rows of document matrix $X$), $m$ being the upper bound on number of documents per query, $U$ being the radius of the Euclidean ball denoting the space of ranking parameters and $R_{\max}$ being the maximum possible relevance value (in practice always $\le$ 5), let $C^{\phi} \in \{C^{sq}, C^{svm}, C^{KL}\}$ be polynomial functions of $R_D, m, U, R_{max}$, where the degrees of the polynomials depend on the surrogate ($\phi_{sq}, \phi_{svm}, \phi_{KL}$). Then we have, 
\begin{equation*}
\ \E_t \left[\|\tilde{z}_t\|^2 \right] \le \dfrac{C^{\phi}}{\gamma} .
\end{equation*}
%where $O(\cdot)$ hides some numerical constants.

\begin{proof}
All our unbiased estimators are of the form $X^{\top} f(s,R,\sigma)$. We will actually get a bound on $f(s,R,\sigma)$ by using Lemma~\ref{normexpr} and $p \to q$ norm relation, to equate out $X$:
\begin{equation*}
\begin{split}
\|\tilde{z}\|_2 & = \|X^{\top} f(s,R,\sigma)\|_2 \le \|X^{\top}\|_{1 \to 2} \|f(s,R,\sigma)\|_1 \\
& \le  R_D   \|f(s,R,\sigma)\|_1
\end{split}
\end{equation*} 
since $R_D \ge \max_{j=1}^m \| X_{j:} \|_2$.

{\bf Squared Loss}: The unbiased estimator of gradient of squared loss, as given in the main text, is:
\begin{equation*}
\tilde{z}= X^{\top} (2 (s  - \dfrac{R_{\sigma(1)} e_{\sigma(1)}}{p(\sigma(1))})) 
\end{equation*} 
where $p(\sigma(1))= \sum_{\pi \in S_m} \mathbb{P} (\pi) \mathbbm{1}(\pi(1)= \sigma(1))$ ($\mathbb{P}= \mathbb{P}_t$ is the distribution at round $t$ as in Alg.~\ref{alg:RTop-kF} )

Now we have:
\begin{equation*}
\|s  - \dfrac{R_{\sigma(1)} e_{\sigma(1)}}{p(\sigma(1))}\|_1 \le m R_D U  + \dfrac{R_{max}}{p(\sigma(1))} \le \dfrac{m R_D U R_{max}}{p(\sigma(1)}
\end{equation*}
Thus, taking expectation w.r.t $\sigma$, we get:
\begin{equation*}
\E_{\sigma} \|\tilde{z}\|^2_2 \le m^2 R_D^4 U^2 R_{max}^2 \E_{\sigma} {\dfrac{1}{p(\sigma(1))^2}}= m^2 R_D^4 U^2 R_{max}^2 \sum_{i=1}^m \dfrac{p(i)}{p^2(i)} 
\end{equation*}
Now, since $p(i) \ge \dfrac{\gamma}{m}$, $\forall \ i$, we get: $\E_{\sigma} \|\tilde{z}\|^2_2 \le $ $ \dfrac{C^{sq}}{\gamma}$, where $C^{sq}= m^4 R_D^4 U^2 R_{max}^2$. 

{\bf RankSVM Surrogate}:  The unbiased estimator of gradient of the RankSVM surrogate, as given in the main text, is:
\begin{equation*}
\tilde{z}=  X^{\top} \left(\dfrac{h_{s,\sigma(1),\sigma(2)}(R_{\sigma(1)}, R_{\sigma(2)}) + h_{s,\sigma(2),\sigma(1)}(R_{\sigma(2)}, R_{\sigma(1)})}{p(\sigma(1),\sigma(2))+ p(\sigma(2),\sigma(1))}\right)
\end{equation*} 
where $h_{s, i,j}(R_i,R_j)= \mathbbm{1}(R_i >R_j) \mathbbm{1}(1+s_j>s_i) (e_j - e_i)$ and $p(\sigma(1), \sigma(2))= \sum\limits_{\pi \in S_m} \mathbb{P} (\pi) \mathbbm{1}(\pi(1)= \sigma(1), \pi(2)= \sigma(2))$ ($\mathbb{P}= \mathbb{P}_t$ as in  Alg.~\ref{alg:RTop-kF})).

Now we have:
\begin{equation*}
\|\dfrac{h_{s,\sigma(1),\sigma(2)}(R_{\sigma(1)}, R_{\sigma(2)}) + h_{s,\sigma(2),\sigma(1)}(R_{\sigma(2)}, R_{\sigma(1)})}{p(\sigma(1),\sigma(2))+ p(\sigma(2),\sigma(1))}\|_1 \le \dfrac{2 }{ p(\sigma(1),\sigma(2)) + p(\sigma(2),\sigma(1))}
\end{equation*}

Thus, taking expectation w.r.t $\sigma$, we get:
\begin{equation*}
\E_{\sigma} \|\tilde{z}\|^2_2 \le 4 R_D^2 \E_{\sigma} \dfrac{1}{(p(\sigma(1), \sigma(2)) + p(\sigma(2),\sigma(1)))^2} \le 4 R_D^2 \sum_{i>j}^m \dfrac{p(i,j) +p(j,i)}{(p(i,j) + p(j,i))^2}
\end{equation*}
Now, since $p(i, j) \ge \dfrac{\gamma}{m^2}$, $\forall \ i,j$, we get: $\E_{\sigma} \|\tilde{z}\|^2_2 \le $ $ \dfrac{C^{svm}}{\gamma}$, where $C^{svm}= O(m^4 R_D^2) $. 

{\bf KL based Surrogate}:  The unbiased estimator of gradient of the KL based surrogate, as given in the main text, is:
\begin{equation*}
\tilde{z}= X^{\top} \left( \dfrac{(\exp(s_{\sigma(1)}) - \exp(R_{\sigma(1)}))e_{\sigma(1)}}{p(\sigma(1))}\right)
\end{equation*}
where $p(\sigma(1))= \sum_{\pi \in S_m} \mathbb{P} (\pi) \mathbbm{1}(\pi(1)= \sigma(1))$ ($\mathbb{P}= \mathbb{P}_t $ as in Alg.~\ref{alg:RTop-kF}) ).

Now we have:
\begin{equation*}
\|  \dfrac{(\exp(s_{\sigma(1)}) - \exp(R_{\sigma(1)}))e_{\sigma(1)}}{p(\sigma(1))}\|_1 \le \dfrac{\exp(R_D U)}{p(\sigma(1))}
\end{equation*}

Thus, taking expectation w.r.t $\sigma$, we get:
\begin{equation*}
\E_{\sigma} \|\tilde{z}\|^2_2 \le R_D^2 \exp(2R_D U) \E_{\sigma} (\dfrac{1}{p(\sigma(1))^2} 
\end{equation*}
Following the same arguement as in squared loss, we get: $\E_{\sigma} \|\tilde{z}\|^2_2 \le $ $ \dfrac{C^{KL}}{\gamma}$, where $C^{KL}= m^2 R_D^2 \exp(2R_X U)$. 

\end{proof}

{\bf Proof of Lemma~\ref{RankSVM}} :

\begin{proof}
Let $m=3$. The term associated with the 1st coordinate of $R$, i.e, $R_1$, in the gradient of RankSVM  is: $\mathbbm{1}(R_1>R_2)\mathbbm{1}(1+s_2>s_1) (e_2-e_1)$ + $\mathbbm{1}(R_2>R_1)\mathbbm{1}(1+s_1>s_2) (e_1-e_2)$ + $\mathbbm{1}(R_1>R_3)\mathbbm{1}(1+s_3>s_1) (e_3-e_1)$ + $\mathbbm{1}(R_3>R_1)\mathbbm{1}(1+s_1>s_3) (e_1-e_3)$.  Now let $s_1=1, s_2=0, s_3=0$. Then the term associated becomes: $\mathbbm{1}(R_2>R_1)(e_1-e_2)$ + $\mathbbm{1}(R_3>R_1)(e_1-e_3)$ = $(\mathbbm{1}(R_2>R_1) + \mathbbm{1}(R_3>R_1))e_1 - \mathbbm{1}(R_2>R_1)e_2 - \mathbbm{1}(R_3>R_1)e_3$. Now, if the gradient can be decomposed over $R_1$, then the term associated with $R_1$ should only be a function of $R_1$. More specifically, $(\mathbbm{1}(R_2>R_1) + \mathbbm{1}(R_3>R_1))$ (the non-zero coefficient of $e_1$, in the term associated with $R_1$) should be a function of only $R_1$. Same for the non-zero coefficients of $e_2$ and $e_3$.  

Now assume that the $(\mathbbm{1}(R_2>R_1) + \mathbbm{1}(R_3>R_1))$ can be expressed as a function of $R_1$ only. Then the difference between the coefficient's values, for the following two cases: $R_1=0, R_2=0, R_3=0$ and $R_1=1, R_2=0,R_3=0$, would be same as the difference between the coefficient's values, for the following two cases: $R_1=0, R_2=1,R_3=1$ and $R_1=1,R_2=1,R_3=1$ (Since the difference would be affected only by change in $R_1$ value). It can be clearly seen that the change in value between the first two cases is: $0-0=0$, while the change in value bertween the second two cases is: $2-0=2$. Thus, we reach a contradiction. 
\end{proof}

{\bf Proof of Lemma~\ref{listnet}} :

\begin{proof}
The term associated with the 1st coordinate of $R$, i.e, $R_1$, in the gradient of ListNet is =  $ \sum_{i=1}^m \left(\dfrac{-\exp(R_i)}{\sum_{j=1}^m \exp(R_j)} + \dfrac{\exp(s_i)}{\sum_{j=1}^m \exp(s_{j})} \right) e_i$

Now, $f(R)= \left(\dfrac{-\exp(R_i)}{\sum_{j=1}^m \exp(R_j)} + \dfrac{\exp(s_i)}{\sum_{j=1}^m \exp(s_{j})} \right)$ is the non-zero coefficient of $e_1$. Now, if $f(R)$ would have only been a function of $R_1$, then $\dfrac{\partial^2 f(R)}{\partial R_i \partial R_{j}}$, $\forall \ j \neq i$ would have been zero. It can be clearly seen this is not the case.

Now, the term associated with $R_1$ and $R_2$, in the gradient of ListNet is same as before, i.e, $ \sum_{i=1}^m \left(\dfrac{-\exp(R_i)}{\sum_{j=1}^m \exp(R_j)} + \dfrac{\exp(s_i)}{\sum_{j=1}^m \exp(s_{j})} \right) e_i$ for both

Now, $f(R)= \left(\dfrac{-\exp(R_i)}{\sum_{j=1}^m \exp(R_j)} + \dfrac{\exp(s_i)}{\sum_{j=1}^m \exp(s_{j})} \right)$ is the non-zero coefficient of $e_1$. Now, if $f(R)$ would have only been a function of $R_1$ and $R_2$, then $\dfrac{\partial^3 f(R)}{\partial R_i \partial R_{j} \partial R_{\ell}}$, $\forall \ell \neq i, \ell \neq j$ would have been zero.  It can be clearly seen this is not the case.

The same arguement can be extended for any $k <m$. 

\end{proof}

{\bf Proof of Theorem.~\ref{impossiblegame}}:

\begin{proof}
We will first fix the setting of the online game. We consider $m=3$ and fixed the document matrix $X \in \mathbb{R}^{3 \times 3}$ to be the identity. At each round of the game, the adversary generates the fixed $X$ and the learner chooses a score vector $s \in \reals^3$. Making the matrix $X$ identity makes the distinction between weight vectors $w$ and scores $s$ irrelevant since $s = Xw = w$. We note that allowing the adversary to vary $X$ over the rounds only makes him more powerful, which can only increase the regret. We also restrict the adversary to choose binary relevance vectors. Once again, allowing adversary to choose multi-graded relevance vectors only makes it more powerful. Thus, in this setting, the adversary can now choose among $2^3=8$ possible relevance vectors.  The learner's action set is infinite, i.e., the learner can choose any score vector $s= Xw= \reals^m$. The loss function $\phi(s,R)$ is any NDCG calibrated surrogate and feedback is the relevance of top-ranked item at each round, where ranking is induced by sorted order (descending) of score vector. We will use $p$ to denote randomized adversary one-short strategies, i.e. distributions over the $8$ possible relevance score vectors. Let $s^*_p = \argmin_{s} \E_{R \sim p} \phi(s,R)$. We note that in the definition of NDCG calibrated surrogates, \cite{ravikumar2011ndcg} assume that the optimal score vector for each distribution over relevance vectors is unique and we subscribe to that assumption. The assumption was taken to avoid some boundary conditions.

It remains to specify the choice of $U$, a bound on the Euclidean norm of the weight vectors (same as score vectors for us right now) that is used to define the best loss in hindsight. It never makes sense for the learner to play anything outside the set $\cup_p s^*_p$ so that we can set $U = \max \{ \|s\|_2 \::\: s \in \cup_p s^*_p \}$.

The paragraph following Lemma 6 of Thm. 3 in \cite{piccolboni2001discrete} gives the main intuition behind the argument the authors developed to prove hopelessness of finite action partial monitoring games. To make our proof self contained, we will explain the intuition in a rigorous way. 

{\bf Key insight}: Two adversary strategies $p,\tilde{p}$ are said to be indistinguishable from the learner's feedback perspective, if for every action of the learner, the probability distribution over the feedbacks received by learner is the same for $p$ and $\tilde{p}$. Now assume that adversary always selects actions according to one of the two such indistinguishable strategies. Thus, the learner will always play one of $s^*_p$ and $s^*_{\tilde{p}}$. Now, let $s^*_p \neq s^*_{\tilde{p}}$. Then, the learner incurs a constant (non-zero) regret on any round where adversary plays according to $p$ and learner plays $s^*_p$, or if the adversary plays according to $\tilde{p}$ and learner plays $s^*_{\tilde{p}}$.   We show that in such a setting, adversary can simply play according to $(p+\tilde{p})/2$ and the learner suffers an expected regret of $\Omega(T)$.

Assume that the adversary selects $\{R_1,\ldots, R_T\}$ from product distribution $\otimes p$. Let the number of times the learner plays $s^*_p$ and $s^*_{\tilde{p}}$ be denoted by random variables $N^{p}_1$ and $N^{p}_2$ respectively, where $N^{p}$ shows the exclusive dependence on $p$. It is always true that $N^p_1 + N^{p}_2= T$. Moreover, let the expected per round regret be $\epsilon_p$ when learner plays $s^*_{\tilde{p}}$ , where the expectation is taken over the randomization of adversary. Now, assume that adversary selects $\{R_1,\ldots, R_T\}$ from product distribution $\otimes \tilde{p}$. The corresponding notations become $N^{\tilde{p}}_1$ and $N^{\tilde{p}}_2$ and $\epsilon_{\tilde{p}}$. Then, 
\[
\E_{(R_1,\ldots,R_T) \sim \otimes p} \E_{(s_1,\ldots,s_T)} [\text{Regret}((s_1,\ldots,s_T),(R_1,\ldots,R_T))] = 0 \cdot \E [N^{p}_1] + \epsilon_{p} \cdot \E [N^{p}_2]
\]
 and 
 \[
 \E_{(R_1,\ldots,R_T) \sim \otimes \tilde{p}} \E_{(s_1,\ldots,s_T)} [\text{Regret}((s_1,\ldots,s_T),(R_1,\ldots,R_T))] = \epsilon_{\tilde{p}} \cdot \E [N^{\tilde{p}}_1] + 0 \cdot \E [N^{\tilde{p}}_2]
\]
  Since $p$ and $\tilde{p}$ are indistinguishable from perspective of learner, $\E [N^{p}_1]= \E[N^{\tilde{p}}_1]= \E [N_1]$ and $\E [N^{p}_2]= \E[N^{\tilde{p}}_2]= \E [N_2]$. That is, the random variable denoting number of times $s^*_p$ is played by learner does not depend on adversary distribution (same for $s^*_{\tilde{p}}$.). Using this fact and averaging the two expectations, we get: 
 \[
 \E_{(R_1,\ldots,R_T) \sim {\frac{\otimes p+\otimes \tilde{p}}{2}}} \E_{(s_1,\ldots,s_T)} [\text{Regret}((s_1,\ldots,s_T),(R_1,\ldots,R_T))] = \frac{\epsilon_{\tilde{p}}}{2} \cdot \E [N_1] + \frac{\epsilon_{p}}{2}  \cdot\E [N_2] \ge \min (\frac{\epsilon_p}{2},\frac{\epsilon_{\tilde{p}}}{2})  \cdot \E [N_1 + N_2]= \epsilon \cdot T
\]

Since $\sup_{R_1,\ldots,R_T} \E [\text{Regret}((s_1,\ldots,s_T),(R_1,\ldots,R_T))] \ge \E_{(R_1,\ldots,R_T) \sim {\frac{\otimes p+ \otimes \tilde{p}}{2}}} \E_{(s_1,\ldots,s_T)} [\text{Regret}((s_1,\ldots,s_T),(R_1,\ldots,R_T))]$, we conclude that for every learner algorithm, adversary has a strategy, s.t. learner suffers an expected regret of $\Omega(T)$.

Now, the thing left to be shown is the existence of two indistinguishable distributions $p$ and $\tilde{p}$, s.t. $s^*_p \neq s^*_{\tilde{p}}$.

{\bf Characterization of indistinguishable strategies in our problem setting}: Two adversary's strategies $p$ and $\tilde{p}$ will be indistinguishable, in our problem setting, if for every score vector $s$, the relevances of the top-ranked item, according to s, are same for relevance vector drawn from $p$ and $\tilde{p}$.  Since relevance vectors are restricted to be binary, mathematically, it means that $\forall s$, $\mathbb{P}_{R \sim p} (R_{\pi_s(1)}=1)= \mathbb{P}_{R \sim \tilde{p}} (R_{\pi_s(1)}=1)$ (actually, we also need $\forall s$, $\mathbb{P}_{R \sim p} (R_{\pi_s(1)}=0)= \mathbb{P}_{R \sim \tilde{p}} (R_{\pi_s(1)}=0)$, but due to the binary nature, $\mathbb{P}_{R \sim p} (R_{\pi_s(1)}=1)= \mathbb{P}_{R \sim \tilde{p}} (R_{\pi_s(1)}=1)$ $\implies$ $\mathbb{P}_{R \sim p} (R_{\pi_s(1)}=0)= \mathbb{P}_{R \sim \tilde{p}} (R_{\pi_s(1)}=0)$). Since the equality has to hold $\forall s$, this implies $\forall j \in [m]$, $\mathbb{P}_{R \sim p} (R_j=1)= \mathbb{P}_{R \sim \tilde{p}} (R_j=1)$ (as every item will be ranked at top by some score vector). Hence, $\forall j \in [m]$, $\E_{R \sim p} [R_j]= \E_{R \sim \tilde{p}} [R_j]$ $\implies$ $\E_{R \sim p} [R]= \E_{R \sim \tilde{p}} [R]$. It can be seen clearly that the chain of implications can be reversed. Hence, $\forall s$, $\mathbb{P}_{R \sim p} (R_{\pi_s(1)}=1)= \mathbb{P}_{R \sim \tilde{p}} (R_{\pi_s(1)}=1)$ $\Longleftrightarrow$ $\E_{R \sim p} [R]= \E_{R \sim \tilde{p}} [R]$.

{\bf Explicit adversary strategies}: Following from the discussion so far and Theorem~\ref{calibration}, if we can show existence of two strategies $p$ and $\tilde{p}$ s.t. $\E_{R \sim p} [R]= \E_{R \sim \tilde{p}} [R]$, but $\argsort \left(\E_{R \sim p} \left[\frac{G({\bf R})}{Z_m(R)}\right]\right) \neq \argsort \left(\E_{R \sim \tilde{p}} \left[\frac{G({\bf R})}{Z_m(R)}\right]\right)$, we are done.

The 8 possible relevance vectors (adversary's actions) are $(R_1,R_2,R_3,R_4,R_5,R_6,R_7,R_8)= (000, 110, 101, 011, 100, 010, 001, 111)$. Let the two probability vectors be: $p= (0.0, 0.1, 0.15, 0.05, 0.2, 0.3,0.2,0.0)$ and $\tilde{p}= (0.0,0.3,0.0,0.0,0.15,0.15,0.4,0.0)$. The data is provided in table format in Table.~\ref{relevanceprobability-table}.  

Under the two distributions, it can be checked that $\E_{R \sim p} [R]= \E_{R \sim \tilde{p}} [R]= (0.45,0.45,0.4)^{\top}$. 

However, $\E_{R \sim p} \left[\frac{G({\bf R})}{Z_m(R)}\right]= (0.3533,0.3920,0.3226)^{\top}$, but $\E_{R \sim \tilde{p}} \left[\frac{G({\bf R})}{Z_m(R)}\right]= (0.3339, 0.3339, 0.4000)^{\top}$. Hence, $\argsort \left(\E_{R \sim p} \left[\frac{G({\bf R})}{Z_m(R)}\right]\right)= [2,1,3]^{\top}$ but $\argsort \left(\E_{R \sim \tilde{p}} \left[\frac{G({\bf R})}{Z_m(R)}\right]\right) \in \{[3,1,2]^{\top}, [3,2,1]^{\top}\}$.

\begin{table}[t] 
\caption{Relevance and probability vectors.} 
\label{relevanceprobability-table}
\begin{center}
\tabcolsep=0.110cm
\begin{tabular}{c c c c c c c c c}  
\hline %inserts double horizontal lines 
\hline $p$ & 0.0 & 0.1 & 0.15 & 0.05 & 0.2 & 0.3 & 0.2 & 0.0\\ [0.4ex]
\hline $\tilde{p}$ & 0.0 & 0.3 & 0.0 & 0.0 & 0.15 & 0.15 & 0.4 & 0.0\\ [0.4ex] 
\hline Rel.& $R_1$ & $R_2$ & $R_3$ & $R_4$ & $R_5$ & $R_6$ & $R_7$ & $R_8$ \\ [0.4ex] 
%heading 
\hline % inserts single horizontal line 
& 0 & 1 & 1 & 0 & 1 & 0 & 0 & 1 \\ % inserting body of the table 
& 0 & 1 & 0 & 1 & 0 & 1 & 0 & 1 \\ 
& 0 & 0 & 1 & 1 & 0 & 0 & 1 & 1  \\ 
% [1ex] % [1ex] adds vertical space 
\hline %inserts single line 
\end{tabular} 
\end{center}
\end{table}

\end{proof}

\end{document}